\documentclass[letterpaper, 10 pt, conference]{ieeeconf}  

\IEEEoverridecommandlockouts                              

\usepackage{graphicx}
\usepackage{subfigure}
\usepackage{amsfonts}
\usepackage{multirow}
\usepackage{tabularx} 
\usepackage{booktabs}  
\usepackage{threeparttable} 
\usepackage{algorithm}  
\usepackage{algpseudocode}  
\usepackage{amsmath}  
\usepackage{bm}

\title{\LARGE \bf
Using Detection, Tracking and Prediction in Visual SLAM to Achieve Real-time Semantic Mapping of Dynamic Scenarios
}

\author{Xingyu Chen$^{1}$, Jianru Xue$^{1,\dagger}$, Jianwu Fang$^{1,2}$, Yuxin Pan$^{1}$ and Nanning Zheng$^{1}$ 
\thanks{*This work is partially supported by NSFC Projects 61751308 and U1713217.}
\thanks{$^{1}$The authors are with the Institute of Artificial Intelligence and Robotics, Xi'an Jiaotong University. Xi'an, P.R. China.}%
\thanks{$^{2}$The author is with the School of Electronic and Control Engineering, Chang'an University. Xi'an, P.R. China.}%
\thanks{$^{\dagger}$Corresponding author's email:{\tt\small jrxue@mail.xjtu.edu.cn}}
}

\begin{document}

\maketitle
\thispagestyle{empty}
\pagestyle{empty}

\begin{abstract}
In this paper, we propose a lightweight system, RDS-SLAM, based on ORB-SLAM2, which can accurately estimate poses and build semantic maps at object level for dynamic scenarios in real time using only one commonly used Intel Core i7 CPU. In RDS-SLAM, three major improvements, as well as major architectural modifications, are proposed to overcome the limitations of ORB-SLAM2. Firstly, it adopts a lightweight object detection neural network in key frames. Secondly, an efficient tracking and prediction mechanism is embedded into the system to remove the feature points belonging to movable objects in all incoming frames. Thirdly, a semantic octree map is built by probabilistic fusion of detection and tracking results, which enables a robot to maintain a semantic description at object level for potential interactions in dynamic scenarios. We evaluate RDS-SLAM in TUM RGB-D dataset, and experimental results show that RDS-SLAM can run with 30.3 ms per frame in dynamic scenarios using only an Intel Core i7 CPU, and achieves comparable accuracy compared with the state-of-the-art SLAM systems which heavily rely on both Intel Core i7 CPUs and powerful GPUs.

\end{abstract}

\section{INTRODUCTION}

Simultaneous Localization and Mapping (SLAM)~\cite{cadena2016past} is an important technique of perception and navigation for intelligent mobile systems, such as robots and autonomous vehicles. Due to the low cost, high resolution, and rich color information of camera, visual SLAM (vSLAM) has become an important research topic over the last years. Some excellent vSLAM systems have been established, such as ORB-SLAM2~\cite{Mur2017ORB}, ElasticFusion~\cite{whelan2016elasticfusion}, RTAB-Map~\cite{RTAB-Map}.

However, classical vSLAM systems commonly assume that scenes are rigid and static, and this assumption leads to frequent failures of vSLAM systems in dynamic scenarios, where there are movable objects, such as people and cars. Even ORB-SLAM2~\cite{Mur2017ORB}, one of the state-of-the-art vSLAM systems, may frequently fail in dynamic scenarios, and can only provide a map with incomplete descriptions. Its localization accuracy is also dramatically degraded. Obviously, these limitations are caused by movable objects in dynamic scenarios.

To overcome the effects of movable objects in dynamic scenarios to vSLAM systems, we propose three major improvements for ORB-SLAM2, and implement a robust and real-time vSLAM framework, RDS-SLAM, for mapping dynamic scenarios. The proposed RDS-SLAM can effectively remove the feature points belonging to movable objects, and build a semantic octree map at object level for complete description of dynamic scenarios. 

\begin{figure}[]
      \centering
      \includegraphics[width=1\linewidth]{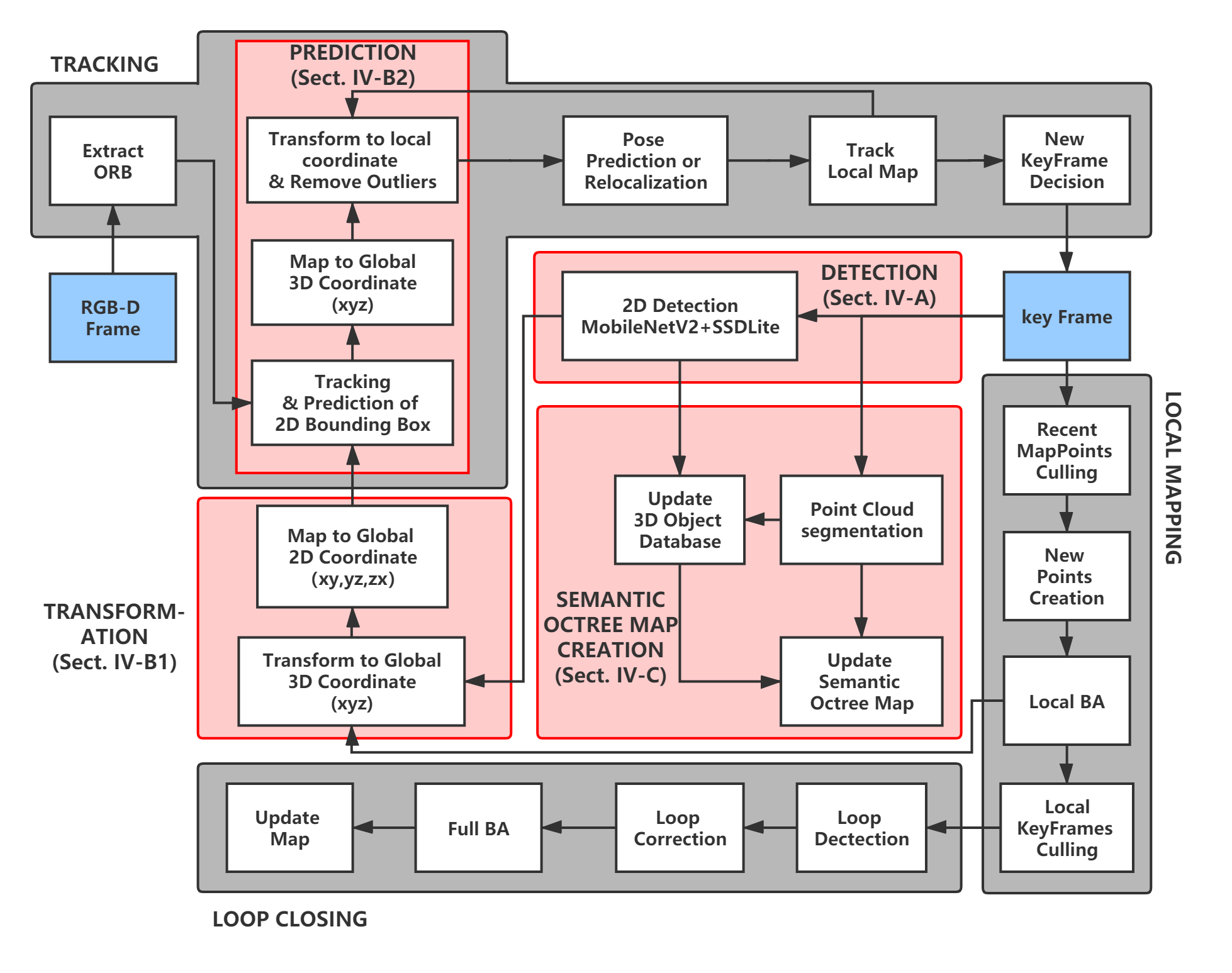}
      \caption{The framework of RDS-SLAM. The threads filled with gray color are original threads of ORB-SLAM2. We denote the improvements over ORB-SLAM2 with red color. The threads in red color are the parallel improvements proposed in this paper. Additionally, a \emph{Prediction} module is inserted into the \emph{Tracking} thread of ORB-SLAM2.
      }
      \label{RDS-SLAM}
   \end{figure}

More specifically, the proposed  improvements, as well as major architectural modifications, are illustrated in Fig.~\ref{RDS-SLAM}. Firstly, we adopt a 2D object detection network as a parallel thread, which is denoted as \emph{Detection} in Fig.~\ref{RDS-SLAM}, and the technical details are presented in Sect.~\ref{subsection1}. Instead of detecting in all frames as other dynamic SLAM systems do, we run it only in key frames to get the 2D movable objects. 

Secondly, we propose an efficient prediction mechanism, which is denoted as \emph{Transformation} and \emph{Prediction} in Fig.~\ref{RDS-SLAM}. We transform the local 2D bounding box to global 3D coordinate and extend the classic local 2D tracking algorithm SORT~\cite{SORT} to global 3D coordinate to track 3D movable objects in key frames, and the constant velocity model is taken to predict other frames (Sect.~\ref{subsection2}). The running time of each frame of the prediction mechanism that we test on Intel i7 CPU is only 5ms.

Finally, we build \emph{Semantic Octree Map Creation} as a parallel thread shown in Fig.~\ref{RDS-SLAM} for both removing dynamic objects and creating a complete semantic map at object level. Instead of raising probability threshold of octree map like other state-of-the-arts do in dynamic scenarios, we use semantic information to distinguish whether the point clouds are movable or not, then insert octree maps with different probabilities to remove the movable object (Sect.~\ref{subsection3}).

The rest of the paper is structured as follows: Section II discusses the related works. Section III presents an overview of RDS-SLAM. Three major improvements are detailed in Section IV, which are followed by experimental results in Section V. Finally, the paper is concluded with discussions and lines for the future in Section VI.
%
%
\section{RELATED WORKS}

There are many excellent vSLAM systems in literature for mapping scenarios~\cite{Mur2017ORB},~\cite{whelan2016elasticfusion},~\cite{RTAB-Map} by using RGB-D data, and a comprehensive survey can be found in~\cite{cadena2016past}. However, they often fail in dynamic scenarios, and this leads to many research works in recent years. In this section, we present a brief survey for these research efforts.

The core idea of improving vSLAM systems is to distinguish the dynamic parts of scenarios. For this purpose, it is straightforward to introduce segmentation~\cite{mccormac2017semanticfusion},~\cite{StaticFusion},~\cite{Co-fusion},~\cite{runz2018maskfusion}. McCormac et al.~\cite{mccormac2017semanticfusion} estimated poses and created a dense map through ElasticFusion~\cite{whelan2016elasticfusion}, then built a single-frame map through a convolutional neural network (CNN) and finally merged two maps to generate a dense semantic map with higher classification accuracy than single-frame CNN. However, it cannot handle the dynamic scenarios. StaticFusion~\cite{StaticFusion}, Co-Fusion~\cite{Co-fusion}, and MaskFusion~\cite{runz2018maskfusion} had been proposed to deal with the dynamic scenarios. They focused on using segmentation information to directly build an accurate dense map that can distinguish the dynamic objects and static scenarios. However, these works have relatively low localization accuracy and heavily rely on intensive computation efficiency.

Among many vSLAM works, ORB-SLAM2~\cite{Mur2017ORB} is widely accepted as the best open source vSLAM system with high localization accuracy and map reusability, but it also fails in dynamic scenarios. The situation has been significantly improved by DynaSLAM~\cite{bescos2018dynaslam} and DS-SLAM~\cite{yu2018ds}, which are two important variants of ORB-SLAM2. To remove the ORB~\cite{ORB} feature points of dynamic objects, DynaSLAM serially added \emph{Mask R-CNN}~\cite{he2017mask}, \emph{Low-Cost Tracking} and \emph{Multi-view Geometry} to the front of ORB-SLAM2 before extracting the ORB feature points. However, since it serially added three modules to the front of ORB-SLAM2, the average time it took per frame using CPU+GPU is about $500$ $ms$. Similar to DynaSLAM, DS-SLAM also serially added \emph{Moving Consistency Check} module and \emph{Remove Outliers} module to the \emph{Tracking} thread of ORB-SLAM2. Different from DynaSLAM, DS-SLAM parallel added \emph{SegNet}~\cite{badrinarayanan2017segnet} thread and \emph{Dense Map Creation} thread to ORB-SLAM2. It finally combines the results of parallel \emph{SegNet} thread and the serial \emph{Moving Consistency Check} module in each frame. Even with such a parallel architecture, its average time of processing a frame using CPU+GPU is about $59.4$ $ms$. In summary, neither DynaSLAM~\cite{bescos2018dynaslam} nor DS-SLAM~\cite{yu2018ds} can work in real time without GPUs, and thus cannot meet with lightweight applications. 

Motivated by the aforementioned works, we propose the real time RDS-SLAM, which can build a complete semantic octree map of dynamic scenario without using GPUs as well as the competitive accuracy compared with DynaSLAM~\cite{bescos2018dynaslam} and DS-SLAM~\cite{yu2018ds}.

\section{SYSTEM OVERVIEW}

We propose a real-time and lightweight RGB-D vSLAM system in dynamic scenarios based on ORB-SLAM2~\cite{Mur2017ORB}. We use object detection and object tracking only in key frames, and use low-cost prediction in other frames to reduce the computational cost, as shown in Fig.~\ref{RDS-SLAM}.

In addition to \emph{Tracking}, \emph{Local Mapping} and \emph{Loop Closing}, three parallel threads of original ORB-SLAM2, we add \emph{Detection}, \emph{Transformation} and \emph{Semantic Octree Map Creation}, three parallel threads into the system. And we also insert a new module named \emph{Prediction} into the \emph{Tracking} thread.

After the processing of \emph{Extract ORB}, \emph{Pose Prediction or Relocalization} and \emph{Track Local Map} in \emph{tracking} thread, ORB-SLAM2 has realized a visual odometry that can estimate the pose transformation between frames in a static scenario. In order to build the map and optimize the pose, ORB-SLAM2 proposes \emph{New KeyFrame Decision} module to select key frames from visual sequence and put them into \emph{Local Mapping} thread. The mechanism of \emph{New KeyFrame Decision} emphasizes that when the scenario changes, a key frame will be inserted after a certain time interval is met, and when the scenario changes quickly, a key frame will be directly inserted regardless of the time interval.

In RDS-SLAM, we believe that detecting objects in all frames without a selection will cost too many computational resources, because the scenario does not always change during localization and mapping of robots. In other words, we should use \emph{Detection} thread only when scenario changes and increase the frequency of detection when scenario changes quickly. Thus we can utilize the mechanism of \emph{New KeyFrame Decision} as the mechanism of \emph{Detection} to realize the adaptive computational resource allocation of \emph{Detection} thread by using \emph{Detection} only in key frames instead of all the frames.

After \emph{New KeyFrame Decision} putting the key frames into the \emph{Local Mapping} thread, ORB-SLAM2 will check the recently added feature points on the map (map points) by \emph{Recent MapPoints Culling} module, as shown in Fig.~\ref{RDS-SLAM}. It emphasizes that if a map point is constructed, it must be observed by the next three key frames. ORB-SLAM2 effectively eliminates the incorrect map points through \emph{Recent MapPoints Culling}, but it cannot effectively remove the map points on movable objects.

On the contrary, in RDS-SLAM, the map points on movable objects of the latest key frame were temporarily built into the map. After RDS-SLAM detecting the latest key frame using an object detection network and putting the results into each of the future frames, the map points on movable objects built by the latest key frame will no longer be observed in the next key frame. Thus RDS-SLAM will remove the map points on movable objects through the principle of \emph{Recent MapPoints Culling}.

\begin{figure}[thpb]
\includegraphics[width=1\linewidth]{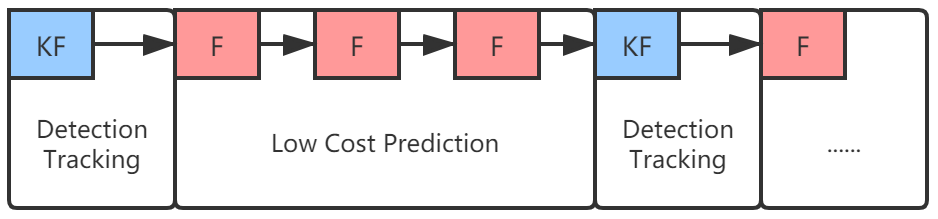}
\caption{The procedure of dealing with movable objects. The boxes with blue color represent key frames (KF), we detect objects in KF. In order to estimate the motion of movable objects, movable objects in KF are being tracked, and then their localizations will be predicted in other frames (F) which are labeled with red color.} \label{2}
\end{figure}

\section{MAJOR IMPROVEMENTS}

For clear description, we elaborate the \emph{Detection} thread, \emph{Transformation} thread, \emph{Prediction} module, and the \emph{Semantic Octree Map Creation} thread of RDS-SLAM in this section.

\subsection{Detection}

We use 2D detection MobileNetv2 SSDLite module in \emph{Detection} thread. The input is RGB-D data of key frame, the output is 2D bounding boxes, depth of the center of bounding boxes, and 20 labels of objects (car, person, tvmonitor, etc.).

We adopt the MobileNetv2 SSDLite~\cite{sandler2018mobilenetv2} object detection network as our detector. The MobileNetV2 SSDLite, as Google's design for mobile, is one of the most efficient object detection networks~\cite{huang2017speed} which could achieve $200$ $ms$ per frame on the Google Pixel 1 phone~\cite{sandler2018mobilenetv2}.

In order to further improve the performance, RDS-SLAM uses NCNN~\cite{NCNN}, the high-performance neural network forward propagation framework of Tencent open API, to accelerate MobileNetV2 SSDLite. We tested it on Ubuntu system, and NCNN MobileNetv2 SSDLite performed on Intel Core i7 CPU to achieve $50$ $ms$ per frame.

\label{subsection1}

\subsection{Transformation and Prediction}
\begin{figure}[thpb]
\includegraphics[width=1\linewidth]{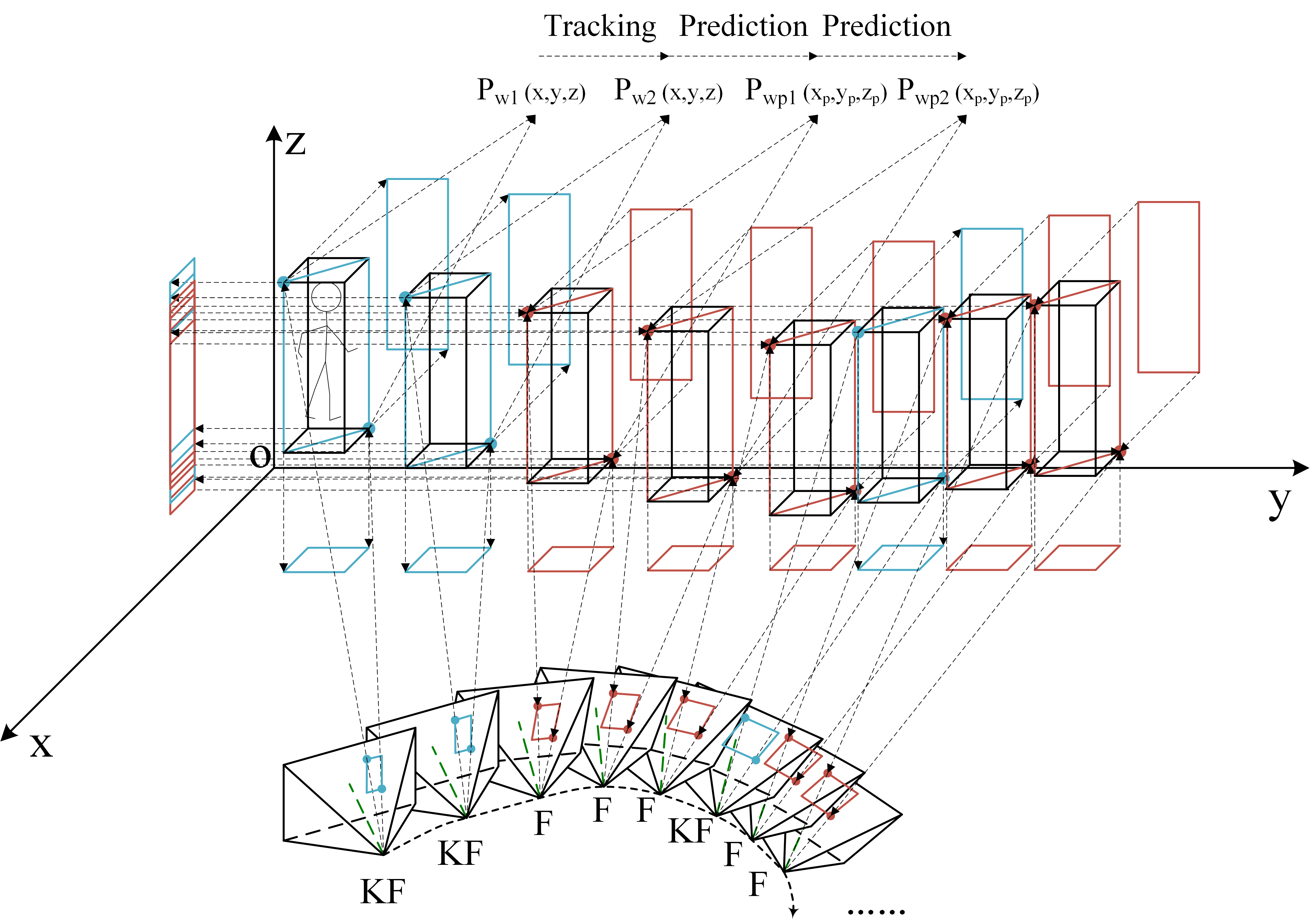}
\caption{An example of \emph{Transformation} and \emph{Prediction}. The quadrangular pyramid represents the pose and image plane of camera. The 3D bounding boxes in blue represent the detected movable objects in key frames (KF). The points on top left and bottom right corners of bounding box are denoted as \bm{$P_{w 1}(x,y,z)$}, \bm{$P_{w 2}(x,y,z)$}, etc. Each 3D bounding box in KF is transformed into the global 2D coordinate planes  ($xOy$, $yOz$, $zOx$), which is then used to predict the 3D bounding boxes in other frames (F). We also use two points to represent a bounding box predicted in F such as \bm{$P_{w p 1}(x_p,y_p,z_P)$}, \bm{$P_{w p  2}(x_p,y_p,z_p)$}, and label the predicted bounding boxes with red color.} \label{RDS-SLAM3D}
\end{figure}

To reach real-time requirements and save computational resources, we use \emph{Detection} only in key frames. In order to deal with other frames, we use \emph{Prediction} module to predict the position of the movable objects, as shown in Fig.~\ref{2}. It means that we will use the movable objects' 2D bounding boxes of latest key frame to predict the movable objects' 2D bounding boxes of current frame, illustrated by Fig.~\ref{RDS-SLAM3D}. 

In order to perform \emph{Transformation} and \emph{Prediction} in real time, we use the depth of the center of bounding box from 2D MobileNetv2 SSDLite module as the depth of the bounding box instead of waiting for \emph{Update 3D Object Database} module. Let \bm{$T_{kcw}$} denote the transformation matrix of the latest key frame, \bm{$T_{cw}$} specify the transformation matrix of the current frame. Notice that we will use the matrix comes from \emph{Tracking} thread as \bm{$T_{kcw}$} before \emph{Local BA} optimization module is complete, and we approximate the transformation matrix of the latest frame as \bm{$T_{cw}$}. Let \bm{$P_{kc}$} denote the bounding box positions matrix of the latest key frame in the camera local coordinate,  and \bm{$P_{c}$} is the bounding box positions matrix of the current frame in the camera local coordinate which can be calculated as:
\begin{equation}
\bm{P_{c}}=\bm{T_{c w}} \times prediction\left(\bm{T_{k c w}^{-1}} \times \bm{P_{k c}}\right),
\end{equation}
where $prediction\left(\bm{P_{w}}\right)$ is an algorithm that we propose to track and predict the change of \bm{$P_{w}$}, and $\bm{P_{w}}$ specifies bounding box positions matrix in the global coordinate. We extend SORT~\cite{SORT} to track the multiple detection results of the key frames by mapping \bm{$P_{w}$} to $xOy$, $yOz$, $zOx$ 2D coordinate planes respectively, where SORT is an algorithm using constant velocity Kalman filter framework~\cite{filtering} and Hungarian algorithm~\cite{Hungarian} to track 2D bounding boxes. Then, the constant velocity model is taken to predict the positions of them in the current frame. After that, we use the information of latest frame to generate the 3D positions of them. Finally, we use Intersection-Over-Union (IOU) to match them to get the prediction result of bounding box positions matrix \bm{$P_{w p}$}. We detail the 3D object prediction in Algorithm~\ref{alg:Prediction}.

\renewcommand{\algorithmicrequire}{\textbf{Input:}}  
\renewcommand{\algorithmicensure}{\textbf{Output:}} 

\begin{algorithm}[htb]\footnotesize
  \caption{ Algorithm of 3D object prediction}  
  \label{alg:Prediction}  
  \begin{algorithmic}[1]  
    \Require  
      The bounding box positions matrix, \bm{$P_w(x,y,z)$}; or $empty$;
    \Ensure  
      Prediction of bounding box positions matrix, \bm{$P_{w p}(x_p,y_p,z_p)$};  
    \State Mapping matrix \bm{$P_w$} to global 2D coordinate planes ($xOy$, $yOz$, $zOx$), as \bm{$P_{x y}(x,y)$}, \bm{$P_{y z}(y,z)$} and \bm{$P_{z x}(x,z)$};
    
    \If {$\textit{input}$ $\neq$ $\textit{empty}$}  
    \State Using SORT which includes constant velocity Kalman filter framework and Hungarian algorithm to track and predict the positions of \bm{$P_{x y}$}, \bm{$P_{y z}$}, \bm{$P_{z x}$} simultaneously and update them;  
    \Else
    \State Using constant velocity model to predict positions of \bm{$P_{x y}$}, \bm{$P_{y z}$}, \bm{$P_{z x}$} in the current frame and update them; 
    \EndIf  
    
    \State Find the max matrix of \bm{$P_{x y}$}, \bm{$P_{y z}$}, \bm{$P_{z x}$}, name it as \bm{$P_{a b}(a,b)$}, and \bm{$P_{b c}(b,c)$}, \bm{$P_{c a}(a,c)$} for others;
    \State Use the information of latest frame to generate \bm{$P_{a b w}(a,b,0)$}, \bm{$P_{b c w}(a_l,b,c)$} and \bm{$P_{c a w}(a,b_l,c)$};
    
    \For{each \bm{$p_{a b w}$} in \bm{$P_{a b w}$}}
        \For{each \bm{$p_{b c w}$}, \bm{$p_{c a w}$} in \bm{$P_{b c w}$}, \bm{$P_{c a w}$}}
            \State Calculate IOU of \bm{$p_{a b w}$} with \bm{$p_{b c w}$}, \bm{$p_{c a w}$} in coordinate $aOb$;
            \State Record the largest IOU and its \bm{$p_{index}$} in \bm{$p_{b c w}$} or \bm{$p_{c a w}$};
        \EndFor
        \State The \bm{$c$} of \bm{$p_{a b w}(a,b,c)$} is equal to the \bm{$c$} of \bm{$p_{index}$};
    \EndFor
    \State $\bm{P_{w p}(x_p,y_p,z_p)}=\bm{P_{a b w}(a,b,c)}$;
  \end{algorithmic}  
\end{algorithm}  

After \emph{Transformation} and \emph{Prediction}, we could use the detection results of the key frame to predict the positions of the movable objects in current frame instead of using object detection within all the frames, as shown in Fig. 4(b). And after that, RDS-SLAM will use the result of prediction to remove the feature points on movable objects in the current frame like Fig. 4(d). So we can see that an artificially created static scenario is provided for the SLAM system.

It is worth noting that the \emph{Prediction} that we serially added in the \emph{Tracking} thread is very lightweight, the running time of each frame of \emph{Prediction} is only $5$ $ms$. Therefore, we achieve the real-time requirements using only CPU.

\begin{figure}[thpb]
\centering
\subfigure[Detection without RDS-SLAM]{
\includegraphics[width=0.46\linewidth]{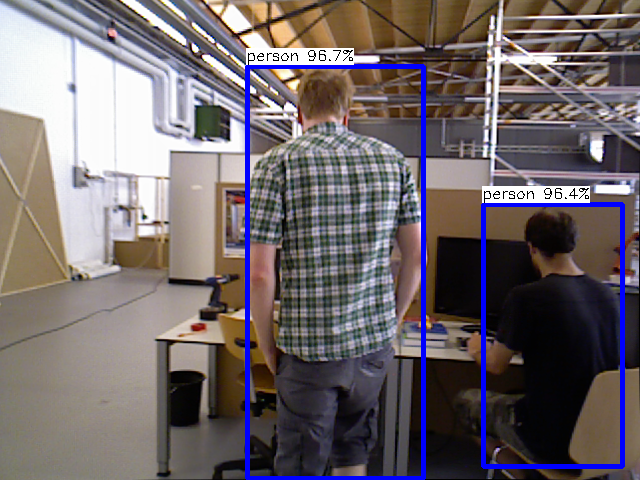}
}
\subfigure[Prediction of RDS-SLAM]{
\includegraphics[width=0.46\linewidth]{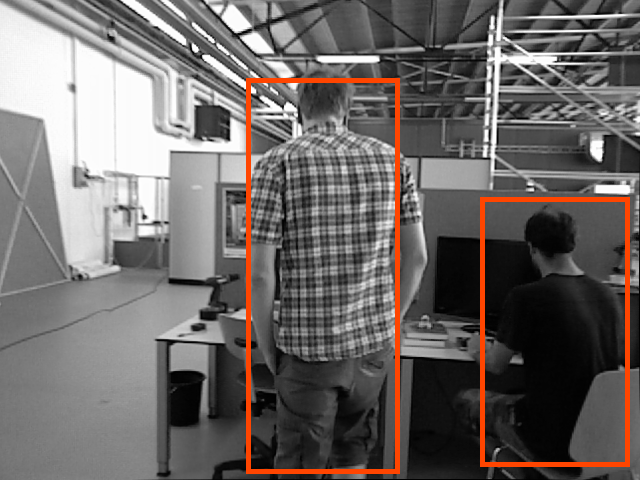}
}
\subfigure[ORB matching of ORB-SLAM2]{
\includegraphics[width=0.46\linewidth]{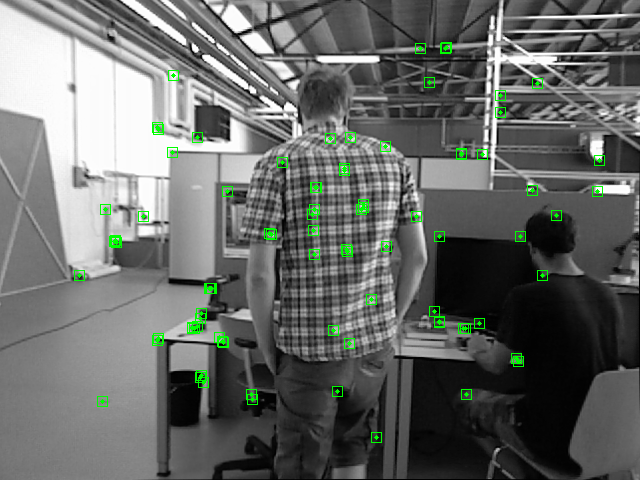}
}
\subfigure[ORB matching of RDS-SLAM]{
\includegraphics[width=0.46\linewidth]{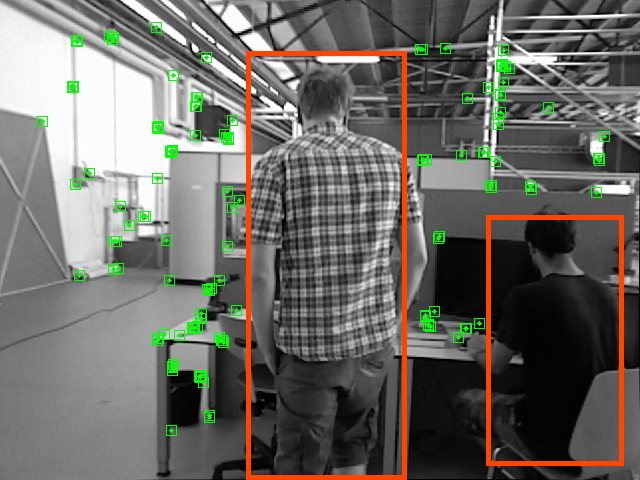}
}
\caption{Test results on \emph{w\_xyz sequence}. Fig. 4(a) shows the test results of the current frame using detection alone without RDS-SLAM. Fig. 4(b) shows the result of the current frame using \emph{Prediction} of RDS-SLAM. We could see that the result of \emph{Prediction} is very close to the result of detecting alone. Fig. 4(c) shows the result of the ORB matching of ORB-SLAM2. Fig. 4(d) shows the result of ORB matching of RDS-SLAM. It is obvious that the result of ORB matching of RDS-SLAM has effectively removed the feature points on the movable objects and got more ORB matching of static scenario.}\label{3}
\vspace{-0.3cm}
\end{figure}
\label{subsection2}

\subsection{Semantic Octree Map Creation}
To provide a complete map description, we build a semantic octree map to show whether a location is occupied, what an occupied point represents for in object level and remove the movable occupied point at the same time. Firstly, we use \emph{Point Cloud Segmentation} module to segment and cluster 3D point cloud to obtain 3D object’s boundary. Then we use \emph{Update 3D Object Database} module to project it to the 2D bounding box to create the 3D bounding box. The \emph{Point Cloud Segmentation} and \emph{Update 3D Object Database} module comes from Ewenwan's open source code~\cite{Ewenwan} which is based on Point Cloud Library (PCL)~\cite{rusu2011point} to slowly build a 3D object database from 3D point cloud and 2D bounding box, so we create the semantic octree map in key frames.

The \emph{Update Semantic Octree Map} module proposed by us will firstly filter the current point cloud and put it into the candidate occupied point cloud. After that, the module will query whether a candidate occupied point is located in the 3D bounding box. If it is true, it will be assigned the corresponding color according to its label. Otherwise, it will retain the original color. In the meanwhile, the module will query whether the candidate occupied point is located in the 3D bounding box of movable objects. If it is, it will be marked as a movable point.

Last but the most important part is updating the probability of occupation of octree map. The octree map~\cite{hornung2013octomap} uses probability to indicate whether a node is occupied, and its occupation probability will be updated when there is a new observation. Assuming $n$ is one of these nodes, then we map the occupancy probability $P(n) \in[0,1]$ of the node $n$ to the logistic regression variable $L(n) \in \mathbb{R}$ and then the update is performed on the space $\mathbb{R}$. The mapping can be calculated as $L(n)=\log \left(\frac{P(n)}{1-P(n)}\right)$.
Assume that the observation at time $T$ is $Z_T$, then the updated formula for occupation probability of the node $n$ can be mapped to:

\begin{equation}
L\left(n | z_{1 : T}\right)=L\left(n | z_{1 : T-1}\right)+L\left(n | z_{T}\right).
\end{equation}
In general, if a node $n$ is inserted by any point at time $T$, $L\left(n | z_{T}\right)=\tau$, and if not, $L\left(n | Z_{T}\right)=0$. The occupation probability threshold is $p$, while the principle that the node $n$ is considered to be occupied is $L\left(n | z_{1 : T}\right)>\log \left(\frac{p}{1-p}\right)$.

Previous SLAM systems with octree map remove moving objects by increasing threshold $p$. These methods can create an octree map without moving objects, but cannot remove potentially movable objects. And after increasing threshold $p$, when the robot moves fast and the observation time for each scenario is limited, the static scenarios that are less observed will also be removed.

To this end, RDS-SLAM removes the movable objects by inserting the points marked as movable or static with different probability. So that it could remove all the movable objects including moving objects and potentially movable objects, and it will not remove the less observed static scenarios at the same time.

In our experiment, we set the occupancy probability threshold to the default $p=0.5$. For the points marked as movable, we set $L\left(n | Z_{T}\right)=-0.41$, for other points, we set $L\left(n | Z_{T}\right)=0.85$, if unobserved, $L\left(n | Z_{T}\right)=0$.

\label{subsection3}

\section{Experimental Results}
\subsection{Dataset, Experimental Setting and Evaluation Metrics}

We implement RDS-SLAM on a Ubuntu operation system running on an Intel i7 CPU without any GPU accelerators. TUM RGB-D dataset~\cite{sturm2012benchmark} is used for evaluation and comparison of RDS-SLAM with other state-of-the-art vSLAM systems. The dataset contains four high-dynamic sequences \emph{w\_xyz}, \emph{w\_static}, \emph{w\_rpy} and \emph{w\_halfsphere} which have two walking men in front of the cameras. It also has two low-dynamic sequences \emph{s\_xyz} and \emph{s\_static} which have two sitting men in front of the cameras. The root mean square error ($RMSE$) of the estimated trajectory with respect to the ground truth is used as the accuracy metric of a vSLAM system~\cite{sturm2012benchmark}. To avoid the impact of non-deterministic nature, we calculate the $RMSE$ of RDS-SLAM with six dynamic sequences, and run the system with each sequence for 10 times. Then we use the median $RMSE$ of each sequence as the accuracy metric of RDS-SLAM. 

Additionally, we also compare the efficiency of RDS-SLAM with the state-of-the-arts in term of computational cost, which is measured with processing time per frame by taking the computing platform into consideration.

\subsection{Analysis and Discussions}
\begin{figure}[t]
\centering
\includegraphics[width=0.9\linewidth]{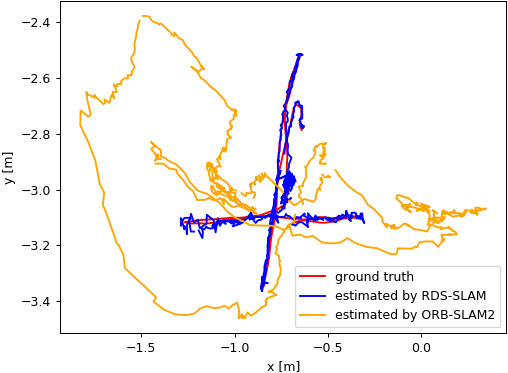}
\caption{Test results on \emph{w\_xyz sequence}. Figure shows the estimated trajectories of ORB-SLAM2 and RDS-SLAM, which are denoted with colors of orange and blue respectively, while the ground truth marked by red color is given.}\label{4}
\end{figure}

For comparison of RDS-SLAM with ORB-SLAM2, we test both of them using the sequence \emph{w\_xyz}. Fig.~\ref{4} shows both the estimated trajectories of ORB-SLAM2 and RDS-SLAM as well as the ground truth. It clearly shows that the estimated trajectory of RDS-SLAM coincided well with the ground truth, while ORB-SLAM2 fails most of the time.

We then compare RDS-SLAM with 7 vSLAM systems~\cite{Mur2017ORB},~\cite{whelan2016elasticfusion},~\cite{StaticFusion},~\cite{Co-fusion},~\cite{runz2018maskfusion},~\cite{bescos2018dynaslam},~\cite{yu2018ds} where  StaticFusion~\cite{StaticFusion}, Co-Fusion~\cite{Co-fusion}, MaskFusion~\cite{runz2018maskfusion}, DynaSLAM~\cite{bescos2018dynaslam} and DS-SLAM~\cite{yu2018ds} are the state-of-the-art dynamic SLAM systems. Table~\ref{tab1} shows experimental results in terms of $RMSE$. Results of RDS-SLAM come from our experiments, others are from reports in~\cite{StaticFusion},~\cite{runz2018maskfusion},~\cite{bescos2018dynaslam},~\cite{yu2018ds}. The two best results are shown in bold in Table~\ref{tab1}. It shows that the accuracy of DynaSLAM, DS-SLAM and RDS-SLAM is higher than other SLAM systems.

\newcommand{\tabincell}[2]{\begin{tabular}{@{}#1@{}}#2\end{tabular}} 
\begin{table*}[ht]  
	\setlength\tabcolsep{4pt}
	\centering  
	\begin{threeparttable}  
		\caption{Comparison of $RMSE(cm)$ and computational cost in dynamic scenarios.}  
		\label{tab1}  
		\begin{tabular}{cccccccccccc}  
			\toprule  
			\multirow{2}{*}{Sequence}&  
			\multicolumn{1}{c}{ORB-SLAM2}& \multicolumn{1}{c}{ElasticFusion}& \multicolumn{1}{c}{Co-Fusion}&\multicolumn{1}{c}{StaticFusion}& \multicolumn{1}{c}{MaskFusion}&\multicolumn{2}{c}{DynaSLAM}&
			\multicolumn{2}{c}{DS-SLAM}&\multicolumn{2}{c}{Ours (RDS-SLAM)}\cr  
			\cmidrule(lr){2-2} \cmidrule(lr){3-3} \cmidrule(lr){4-4} \cmidrule(lr){5-5} \cmidrule(lr){6-6}\cmidrule(lr){7-8} \cmidrule(lr){9-10} \cmidrule(lr){11-12}
			&RMSE&RMSE&RMSE&RMSE&RMSE&RMSE&Cost&RMSE&Cost&RMSE&Cost\cr 
			\midrule  
			w\_xyz& 45.9 & 90.6 & 69.6 &12.7& 10.4 & \textbf{1.5} & \multirow{6}{*}{\tabincell{c}{$\approx$ 500 ms\\on CPU +\\M40 GPU}} &2.5& \multirow{6}{*}{\textbf{\tabincell{c}{59.4 ms\\on i7 CPU+\\P4000 GPU}}} & \textbf{1.7} & \multirow{6}{*}{\textbf{\tabincell{c}{30.3 ms\\on i7 CPU}}}      \cr  
			w\_static& 9.0 & 29.3 & 55.1 & 1.4 & 3.5 & \textbf{0.6} & & \textbf{0.8} & & 0.9 & \cr  
			w\_rpy&    66.2 & -   & -    & -   & -   & \textbf{3.5} & & 44.4& & \textbf{3.9} & \cr  
			w\_half&   35.1 & 63.8 & 80.3 & 39.1 & 10.6 & \textbf{2.5} & & \textbf{3.0} & & 3.1 & \cr  
			s\_static& 0.9 & \textbf{0.8} & 1.1 & 1.3 & 2.1 & - & & \textbf{0.7} & & \textbf{0.8} & \cr  
			s\_xyz& \textbf{0.9} & 2.2 & 2.7 & 4.0 & 3.1 & 1.5& & - & & \textbf{1.1} & \cr  
			\bottomrule  
		\end{tabular}  
	\end{threeparttable}  
\end{table*}  

We also compare the efficiency of RDS-SLAM with DynaSLAM and DS-SLAM since they have comparable localization accuracy and have higher accuracy than other systems relying on powerful GPUs. As shown in Table~\ref{tab1}, both DynaSLAM and DS-SLAM need CPUs accelerated with GPUs to achieve the speed of $500$ $ms$ and $59.4$ $ms$ per frame, while RDS-SLAM can achieve a speed of $30.3$ $ms$ per frame only with an Intel i7 CPU.

\begin{table}[ht]\scriptsize
	\setlength\tabcolsep{3pt}
	\centering  
	\begin{threeparttable}  
		\caption{Computational cost analysis.}
		\label{tab2}  
		\begin{tabular}{ccccc}  
			\toprule  
			Framework&Platform&SITC&PITC&ATCPF\cr  
			\midrule  
			DynaSLAM&\tabincell{c}{CPU +\\M40 GPU}&\tabincell{c}{LC Tracking 1.64 ms\\MV Geometry 285 ms\\Mask R-CNN 195 ms}&0&$\approx$ 500 ms \cr  
			DS-SLAM&\tabincell{c}{i7 CPU +\\P4000 GPU}&\tabincell{c}{MC Check 29.5 ms}&\tabincell{c}{SegNet 37.6ms}&59.4 ms \cr  
			RDS-SLAM&\textbf{i7 CPU}&\tabincell{c}{Prediction $\approx$ 5 ms} &\tabincell{c}{MV2SSD 53 ms}&\textbf{30.3 ms}\cr  
			ORB-SLAM2&\textbf{i7 CPU}&0&0&\textbf{25.6 ms}\cr  
			\bottomrule  
		\end{tabular}
	\end{threeparttable}  
\end{table}  

Furthermore, we analyze the computational cost of the top 3 accurate vSLAM systems in details, since they all are improved from ORB-SLAM2. As shown in Table~\ref{tab2}, where SITC denotes the serial increase in time consumption, PITC specifies the parallel increase in time consumption, and ATCPF is the average time consumption per frame. SITC is the key point to determining whether an improved system can run in real time. In DynaSLAM, it serially added \emph{Low-Cost Tracking}, \emph{Multi-view Geometry} and \emph{Mask R-CNN} to improve ORB-SLAM2, which causes an increment of $500$ $ms$ of ATCPF. In DS-SLAM, it introduces \emph{SegNet} in parallel, but each current frame needed to wait for the result of \emph{SegNet}. In the meanwhile, DS-SLAM serially added the \emph{Moving Consistency Check} into ORB-SLAM2. They cause an increment of 30ms of ATCPF. Different from detecting objects in all frames, RDS-SLAM uses the detection results of key frames to predict the positions of movable objects in other frames. Table~\ref{tab2} shows that RDS-SLAM adds MobileNetV2 SSDLite in parallel, and each current frame uses the output from the lightweight \emph{Prediction} without waiting, so the ATCPF of RDS-SLAM is only increased by about 5ms.

Fig. 6 shows the semantic octree maps built by RDS-SLAM. Firstly, we test it in \emph{w\_xyz} sequence. Fig. 6(a) shows the test result of the octree map without semantic association. Fig. 6(b) demonstrates the result of the octree map with semantic association. It is obvious that the octree map with semantic association effectively removes movable objects. Fig.6(c) shows the semantic octree map we build for the \emph{room} sequence comes from TUM RGB-D dataset [23], and it demonstrates that both the geometric information and semantic information are completely presented in the map.

\section{Conclusions}

In this paper, we present an efficient RDS-SLAM system, which is a lightweight visual semantic SLAM system for dynamic scenarios. It runs well with an Intel i7 CPU in real time without using any GPU accelerator. RDS-SLAM is an improved variant of ORB-SLAM2 by the adoption of \emph{Detection}, \emph{Transformation}, \emph{Prediction} and \emph{Semantic Octree Map Creation}, which efficiently removes the feature points belonging to movable objects and builds an accurate semantic octree map at object level for dynamic scenarios. 

The efficiency of RDS-SLAM was validated with the TUM RGB-D dataset. We also compare it with the state-of-the-art vSLAM systems. Experimental results show that RDS-SLAM can run with $30.3$ $ms$ per frame using only an Intel i7 CPU and reach the competitive performance to the state-of-the-art SLAM systems in dynamic scenarios. 

Future extensions of this work might be using semantic description to clarify ambiguity in corresponding feature points, and exploring geometric structures to handle high dynamic scenarios.

\begin{figure}[t]
\centering
\subfigure[Without semantic association]{
\includegraphics[width=0.46\linewidth]{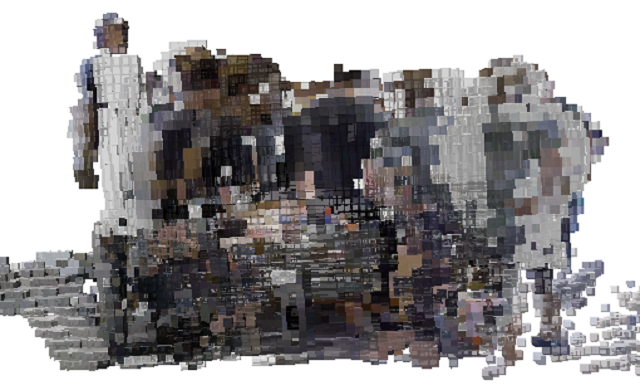}
}
\subfigure[With semantic association]{
\includegraphics[width=0.46\linewidth]{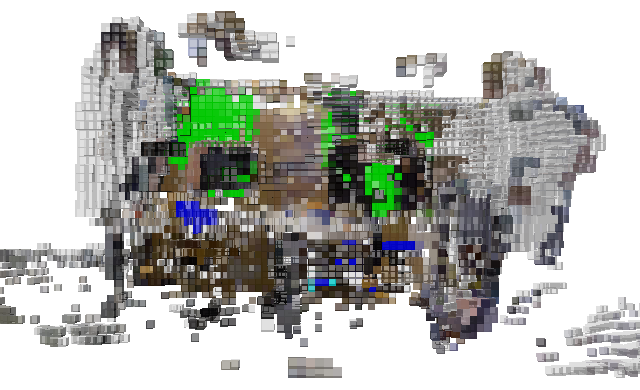}
}
\subfigure[Map with semantic association in \emph{room} sequence.]{
\includegraphics[width=0.82\linewidth]{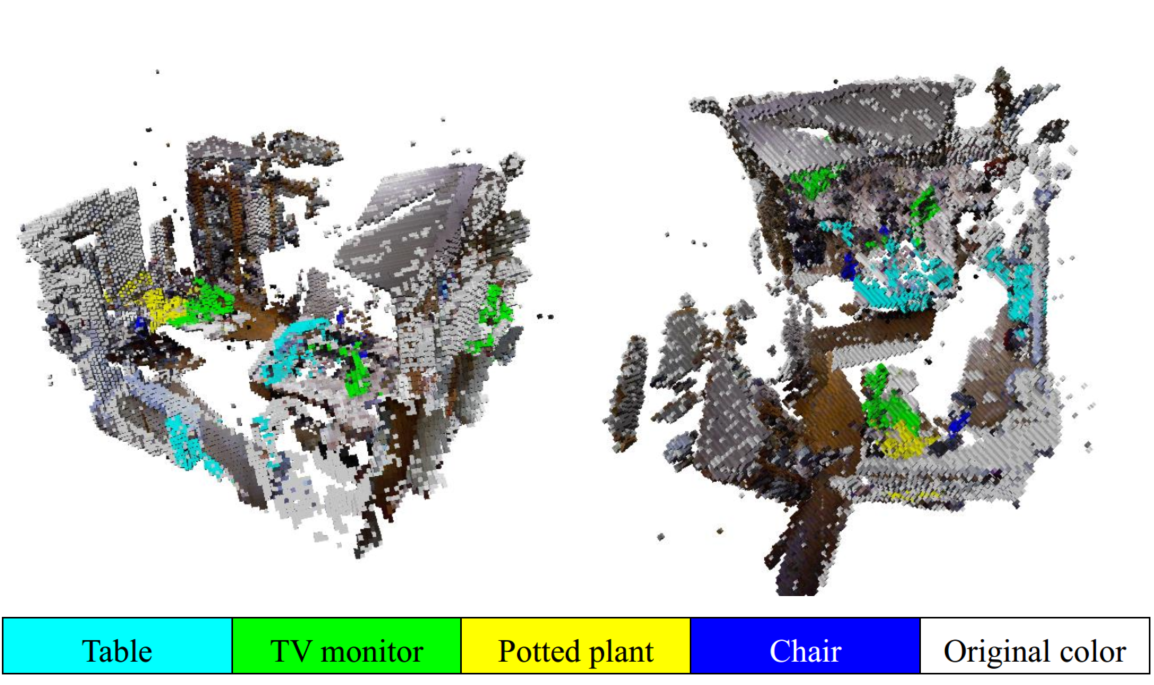}
}
\caption{Octree maps built by RDS-SLAM.}\label{5}
\vspace{-0.4cm}
\end{figure}

{\small{
\bibliographystyle{IEEEtran}
\bibliography{cite}}
}

\end{document}